# An improved sex-specific and age-dependent classification model for Parkinson's diagnosis using handwriting measurement

Ujjwal Gupta[1*], Hritik Bansal[2*], Deepak Joshi[3,4**]

*Equal Contribution   ** Corresponding Author

*Abstract*—
**Background and Objectives:** Diagnosis of Parkinson's with higher accuracy is always desirable to slow down the progression of the disease and improved quality of life. There are evidences of inherent neurological differences between male and females as well as between elderly and adults. However, the potential of such gender and age infomration have not been exploited yet for Parkinson's identification.
**Methods:** In this paper, we develop a sex-specific and age-dependent classification method to diagnose the Parkinson's disease using the online handwriting recorded from individuals with Parkinson's (n = 37; m/f-19/18;age-69.3±10.9yrs) and healthy controls (n = 38; m/f-20/18;age-62.4±11.3yrs). A support vector machine ranking method is used to present the features specific to their dominance in sex and age group for Parkinson's diagnosis.
**Results:** The sex-specific and age-dependent classifier was observed significantly outperforming the generalized classifier. An improved accuracy of 83.75% (SD = 1.63) with the female-specific classifier, and 79.55% (SD = 1.58) with the old-age dependent classifier was observed in comparison to 75.76% (SD = 1.17) accuracy with the generalized classifier.
**Conclusions:** Combining the age and sex information proved to be encouraging in classification. A distinct set of features were observed to be dominating for higher classification accuracy in a different category of classification.

## I. INTRODUCTION

Parkinson's disease (PD) is a progressive, complex neurodegenerative disorder reflecting tremor and loss of postural reflexes. An estimated 10 million people in the world (i.e., approximately 0.3% of the world population) are found to be affected with PD [1], making it second in the list of most common neurodegenerative disorders [2]. In the United States alone, one million people are diagnosed with Parkinson each year [3] with an economic burden of more than $14.4 billion [4]. Diagnosis of Parkinson's with higher accuracy is always desirable to slow down the progression of the disease and improved quality of life.

Structural imaging modalities, such as computerized tomography (CT) and magnetic resonance imaging (MRI), have a limited role in diagnosing PD. Imaging-based diagnosis is expensive and requires specialized skills to operate, which make them impractical for continuous monitoring. Further, the increased iron concentration in the substantia nigra causes decreased signal intensity on T2 weighted images, but these changes are not sufficient to reliably distinguish PD patients from healthy controls [5] and therefore may lead to misdiagnosis. A detailed review of imaging modalities for Parkinson's diagnosis can be found elsewhere [6]. Such misdiagnosis put those patients on wrong drugs and delays the correct treatment. Another misdiagnosis occurs when movement-related information is utilized for Parkinson's diagnosis. For example, Parkinson's is misdiagnosed as a stroke by physicians who not normally see Parkinson's patients [7]. It is due to overlapping movement syndromes and therefore requires a specialist in movement disorders. The accurate determination of PD, however, is vital for patient counseling and clinical research purposes. Early intervention with exercise after a precise judgment of PD can prevent falls [8] and improve quality of life with a reduced cost of care [9]. In addition to correct diagnosis, it is also desirable that the determination method is quick, low-cost, and can be easily operated without specific skills. MRI and CT are expensive, time-consuming, with technical expertise to manage and therefore create scope for an alternate method of diagnosis. Among the current alternative methods, the most popular are wearable sensor-based gait analysis [10,11,12,13] and speech analysis [14,15,16,17,18].

Despite their simplicity in diagnosis method in Parkinson's identification, both speech and gait analyses suffer from some limitations. While speech recording requires high-quality recording with no background noise, the gait monitoring requires specialized instrumentation with enough space to walk. Further, the fear of fall during walking in Parkinson's disease limits the use of gait analysis in Parkinson's disease identification [19]. Micrographia refers to abnormally small and cramped handwriting and is well documented to be associated with Parkinson's disease [20,21,22,23]. Handwriting eliminates the need of noise-free environment and also the gait-related difficulties in measurement. It has also proved to be a potential marker in the diagnosis of Parkinson's disease [24].

The main contributions of the present work are twofold. First, we show an improved Parkinson's diagnosis using sex and age-based classification model. Second, we discuss the capacity of each individual handwriting task by analyzing task-specific features relevant for PD classification.

The rest of the paper is organized as follows. Section II discusses the related work in this area. Section III briefly describes the PaHaW dataset. Section IV, V focuses on the feature extraction and selection pipeline. Section VI reports the results obtained. Section VII provides a critical analysis of the results. Sectionz VIII concludes the work.

## II. RELATED WORK

There has been prior work leveraging handwriting as a biomarker for PD classification. [24, 39, 40] uses kinematic, entropic and energetic features for in-air as well on-surface online handwriting with SVM classifier. [49] extends the feature space by extending the velocity-based signal features. [51] deploys deep learning models – ImageNet, AlexNet for feature extraction and classification respectively. [50] uses a genetic programming approach with kinematic features to overcome the "black box" approach of Artificial Neural Nets(ANNs) and SVMs. However, the proposed system adds sex-specific and age-dependent distinction to understand the impact of this prior imposition on the classification performance.

Sex differences are prominent in Parkinson's disease – 1) higher incidence and prevalence in men, 2) age at disease onset in women is later and 3) higher age and disease duration in women at the time of first Unified Parkinson's Disease Rating Scale (UPDRS-III). [25, 26, 27, 28]. One possible source of male-female differences in the clinical and cognitive characteristics of PD is reported as an effect of estrogen on dopaminergic neurons and pathways in the brain [29, 30, 31, 32]. Previously, it has been shown that the genes expression profiles and survival adaptive process in substantia nigra (SNc) dopaminergic (DA) neurons have different mechanism in male and females [50, 51], which suggest sex-specific nature of the Parkinson's disease as well as the treatment. Similar to sex differences, the age group (middle and old), specific differences are also reported in the literature [33, 34, 35]. For example, for a comparable length of Parkinson's disease duration, the total Unified Parkinson's Disease Rating Scale (UPDRS) motor score is significantly higher in those with old-age PD onset than in those with middle-age onset [36]. We are therefore motivated to introduce sex and age information to the classifier in anticipation of improved classification.

The present work is an extension of [24,37,38] towards the utilization of prior age-and-sex information in anticipation of improved diagnosis of Parkinson's disease. Similar to [49], we also investigate the predictive potential of each task for PD classification. Moreover, our proposed framework of sex/age based distinction can be adapted to other PD detection systems as well.

III. DATA DESCRIPTION

The Parkinson's Disease Handwriting Database (PaHaW) was used in the present work. It consists of multiple handwriting samples from 37 individuals with Parkinson's disease (19 male and 18 female; age – 69.3 ± 10.9 years) and 38 sex and age controls (20 male and 18 female; age – 62.4 ± 11.3 years). In the entire database, the age of 41 subjects was in the range 65 – 92 years while 34 participants were in the range of 36 – 64 years. Each person performed seven writing tasks (aka Task 1-7) in the Czech language, as shown in Fig.1. The writing tasks involved writing cursive letters or bi/tri-grams of letters (Task 1-3), one long stroke writing (Task 4-6), and a longer sentence to capture fatigue effect (Task 7). A digitizing Tablet Intuos 4M (Wacom Technology) was used to acquire the handwritten signals characterized by the seven dynamic features: 1) x-coordinate, 2) y-coordinate, 3) timestamp, 4) button status, 5) pressure, 6) tilt, and 7) elevation. Button status is a binary variable which facilitates the segmentation of on-air and on-surface strokes. Full feedback of the writing during experiments was provided to the participants. A detailed analysis of experiment protocol and data collection technique can be found in [37, 38, 39].

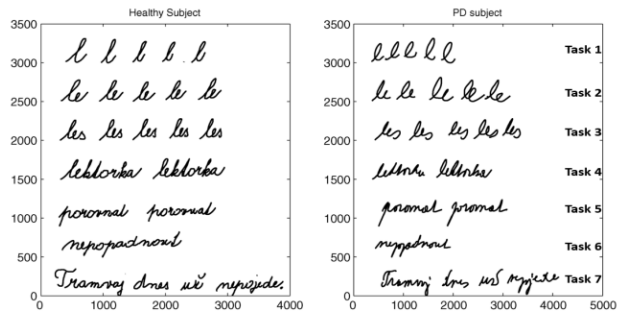

Fig.1 Handwriting sample of healthy and PD subject
(Image Courtesy: [37])

IV. FEATURE EXTRACTION

The features extracted from the handwriting signals are mainly categorized under Kinematic, Entropic, and Energetic features, as illustrated in Table 1. Detailed description and definitions of the features can be found in [37,38,39]. Those features which resulted into a vector under any category were further processed by statistical functions like mean, standard deviation, second-order, third-order moments, robust range, and percentiles to get the entire input feature space (~ 300 features per task). Extraction of Kinematic, Energetic and Entropic features facilitated a broader scope to explore specific handwriting features of the considered classes, both sex, and age-based. All features were considered for both in-air and on-surface movement during classification. Although on-surface kinematics are more appealing for classification, recent work has demonstrated the potential of in-air handwriting movements in Parkinson's identification [40]. It is important to note that features might not be independent like Horizontal and Vertical direction motion is considered as separate features. The absence of directional components of Stroke speed is an indication of its scalar nature. Hidden complexities in the handwriting were modeled using Entropic and energy in the features.

Empirical Mode Decomposition is an important signal processing technique in which a non-linear time-series signal is decomposed into various components (Intrinsic Mode Functions). Entropic and Energetic features for IMFs are also calculated (as presented in [37]). Intrinsic Mode Functions thus derived are used to calculate Intrinsic Conventional Energy and Intrinsic Taeger-Kaiser Energy as presented in Table 1.

V. FEATURE SELECTION AND CLASSIFICATION

To show the impact of prior knowledge about sex and age in Parkinson's classification improvement, the present work developed four schemes of classification as follows: 1) *Generalized Classifier* - Under this scheme, the classifier is trained with no prior age or sex information, 2) *Sex-Specific Classifier* – Under this scheme two classifiers are trained, one

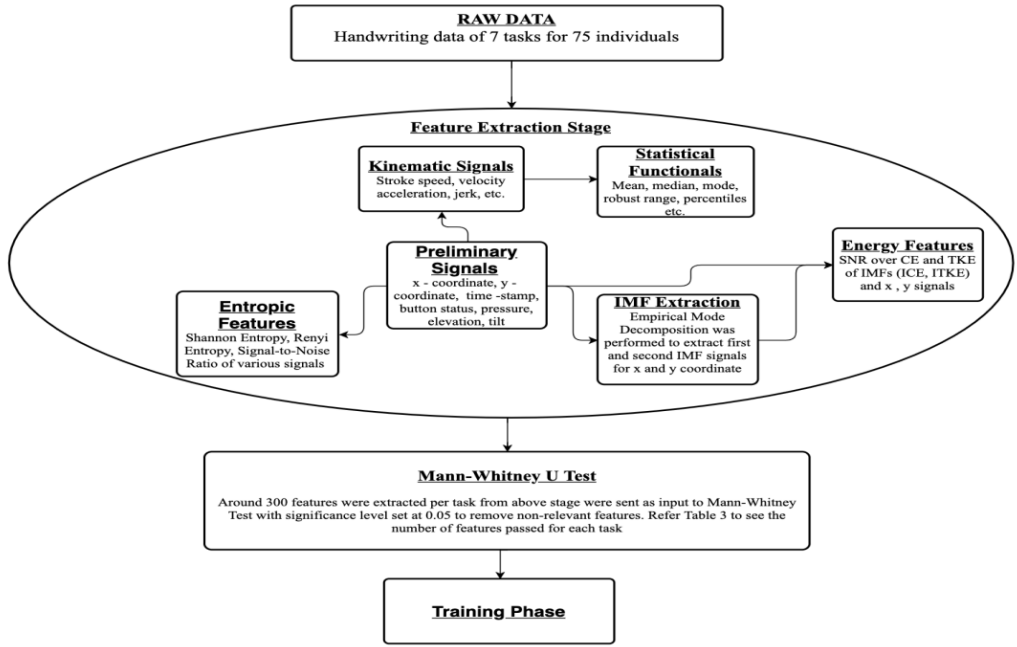
Fig.2 Flowchart for preparing the data for training phase

only with female subjects and other only with male subjects for Parkinson vs. control, 3) *Age-Dependent Classifier* – Under this scheme, two classifiers are trained for Parkinson vs. control, one with old-aged participants (Age>=65 years) and other with young subjects (Age <65 Years), and 4) *Age and sex Dependent Classifier* – Under this scheme four classifiers namely young male, young female, old male, and old female are trained with prior knowledge of age and sex.

TABLE 1
FEATURES EXTRACTED FROM HANDWRITING SIGNALS

| Kinematic Signals | Entropic | Energetic Features |
|---|---|---|
| Strokespeed/Velocity(vel.) | Shannon | X/Y-coordinate |
| Acceleration(acc.)/Jerk | X/Y-coordinate | *Signal to Noise* Ratio for |
| Horizontal Vel./Acc./Jerk | | Conventional |
| Vertical Vel./Acc./Jerk | Second Order Renyi | Energy(CE)/Taeger-Kaiser |
| | X/Y-coordinate | Energy(TKE)/Intrinsic |
| Number of changes in Vel. | | CE/Intrinsic TKE |
| Direction(NCV) | Third Order Renyi | |
| Number of changes in Acc. | X/Y-coordinate | |
| Direction(NCA) | | |
| Pressure Rate | | |

The demographics of individuals with Parkinson's and healthy controls in each scheme of classification are shown in Table 2. The original dataset was split on the basis of various categories-Sex, Age and Sex-Age. It has to be noted that as we develop more groups the amount of data nearly goes half (in Sex-specific and Age-Dependent) and one fourth (in Age and Sex Dependent Classifier) for training a classifier in comparison to a generalized classifier, as shown in Table 2. This distribution of samples in more number of classes causes class imbalance problem when dealing with Age and Sex-Age category classification, which is discussed in the Discussion section of the manuscript. Feature Selection was made in two steps for each class/category specific dataset. Firstly, Mann-Whitney U-Test, as a measure of mutual information, was performed to reduce the dimensionality of input space. Secondly, Support Vector Machines (SVM) ranking method presented in [37] was used for the further selection procedure. Both steps are discussed in brief as in the following subsections and are depicted in the flowchart in Fig.2 and Fig.3. The feature selection was performed for the entire dataset, i.e. training and testing, both.

*A. Mann-Whitney U Test*

To reduce the dimensionality of input data (~300 features for each task) and remove the non-relevant features, the first stage was a statistical analysis using the Mann-Whitney U test performed in MATLAB. The Mann-Whitney U test is a nonparametric statistical test used to assess whether two independent groups are significantly different from each other for a given feature. The only features that passed the Mann-Whitney U-test with a significance level ($p<0.05$) were considered for a ranking using a support vector machine (SVM) ranking [40].

*B. Support Vector Machines (SVM) Ranking Method*

The effectiveness of the selected subset of features in classifying PD and non-PD subjects was evaluated using nonlinear (RBF kernel) SVM. The underlying idea of SVM classifiers is to calculate a maximal margin hyperplane separating two classes of the data.

To learn nonlinearly separable functions, the data are implicitly mapped via nonlinear mapping $\varphi(x)$ to a higher dimensional space employing a kernel function, where a separating hyperplane is found [37]. The relation gives the equation of the hyperplane separating two differential classes

$$y(x) = \sum_{j=1}^{k} w_j \, \varphi_j(x) + w_0 \qquad (1)$$

Where $W = [w_1, w_2, .., w_k]$ is the weight vector k dimensional weight vector.

TABLE 2
DISTRIBUTION OF PATIENTS AND HEALTHY CONTROLS FOR VARIOUS CATEGORIES

| Category | | Number of Patients (PD) | Number of Healthy Controls(HC) |
|---|---|---|---|
| All Subjects | Combined | 37 | 38 |
| Sex | Male (65.35±12.97 years) | 19 | 20 |
| | Female (66.33±10.16 years) | 18 | 18 |
| Age | Subjects with Age < 65 (Young) [18 Males,16 Females] | 11 | 23 |
| | Subjects with Age >= 65 (Old) [21 Males,20 Females] | 26 | 15 |
| Sex-Age | Young Male | 7 | 11 |
| | Old Male | 12 | 9 |
| | Young Female | 4 | 12 |
| | Old Female | 14 | 6 |

New samples are classified according to the side of the hyperplane they belong to. We used the Radial Basis Function (RBF) Kernel. It is defined as

$$K(x, y) = e^{\frac{-||x-y||^2}{2z^2}} \quad (2)$$

Where z controls the width of the RBF function. Python *scikit learn* library was used to implement SVM in our model. We used C= [0.001, 0.003, 0.01, 0.03, 0.1, 0.3, 1, 3, 10, 30, 100, 300, 1000] grid for tuning slack parameter which is inversely related to the extent of regularization and z = [0.03, 0.06, 0.12, 0.25, 0.5, 1, 2, 4, 8, 16, 32]. To investigate the sensitivity of the model for both the number of features added and the order in which the features are added, the following approaches were employed:

1. *Random order approach:* The features are added in random order.
2. *Descending order approach:* The features are arranged in descending order of their individual accuracies and added in this order, as suggested in [37].

In each repetition, the original dataset was randomly permuted, followed by 80:20 train and test split. The best model that we get after performing stratified ten-fold cross-validation is used to determine the accuracy value over the test split. Finally, the accuracy values are averaged over fifty epochs, which are reported as the classification performance values.

Features were normalized to zero mean and standard deviation of one before feeding them to the input of classifier. We define the capacity of a handwriting task for a particular Class (Combined/ Male/ Female/ Young/ Old) as the highest classification accuracy (using SVM) obtained by a feature corresponding to that task.

The features were arranged in the order of their individual classification accuracy. These arranged features were used in the final classifier as illustrated in Flowchart shown in Fig.3. We obtained the classification performance of each feature for each task. These are added and fed to SVM classifier in random or descending order until the maximum classification accuracy was achieved.

*C. Classification Performance Parameters*

The classification performance was determined by the computation of accuracy, precision, and recall. The accuracy ($P_{acc}$), precision ($P_{pre}$) and recall ($P_{rec}$) and are defined as

$$P_{acc} = \frac{TP+TN}{TP+FP+TN+FN} * 100\%$$
$$P_{pre} = \frac{TP}{TP+FP} * 100\%$$
$$P_{rec} = \frac{TP}{TP+FN} * 100\%$$

where true positive (TP) and false positive (FP) represent the number of correctly classified PD subjects and the number of actually healthy subjects diagnosed as PD, respectively. Similarly, true negative (TN) and false negative (FN) represents the total number of correctly classified healthy controls, and the PD patients incorrectly classified as healthy controls, respectively. To evaluate the performance of the model on classes with skewed data, precision and recall are more reliable performance parameter than accuracy. Hence, precision and recall values for young and old classes should provide insight into the model's performance.

VI. RESULTS

Results, in general, indicated that specific handwriting tasks and corresponding features are more likely to be essential for the classification of PD than others depending on their class (Male/Female, Old/Young). Removing redundant features reduced the dimensionality of input space, causing faster learning without loss of accuracy.

TABLE 3
MANN-WHITNEY U TEST RESULTS

| Task Number | Description of Task | # features passed (*Combined*) | # features passed (*Male*) | # features passed (*Female*) | # features passed (*Old*) | # features passed (*Young*) |
|---|---|---|---|---|---|---|
| Task 1 | Cursive letters | 82 | 79 | 15 | 94 | 2 |
| Task 2 | | 84 | 70 | 21 | 119 | 3 |
| Task 3 | | 2 | 3 | 7 | 2 | 3 |
| Task 4 | One long stroke | 53 | 1 | 51 | 12 | 1 |
| Task 5 | | 26 | 9 | 92 | 2 | 1 |
| Task 6 | | 47 | 6 | 20 | 30 | 13 |
| Task 7 | A longer sentence | 177 | 9 | 215 | 216 | 23 |

TABLE 5
CAPACITY OF EACH TASK ACROSS DIFFERENT CLASS

|  | Feature In *Combined* Class (Best Accuracy (%)) | Feature in *Male* Class (Best Accuracy (%)) | Feature in *Female* Class (Best Accuracy (%)) | Feature In *Young* Class (Best Accuracy (%)) | Feature In *Old* Class (Best Accuracy (%)) | Feature in *Old Female* Class (Best Accuracy (%)) | Feature in *Young Female* Class (Best Accuracy (%)) | Feature In *Old Male* Class (Best Accuracy (%)) | Feature In *Young Male* Class (Best Accuracy (%)) |
|---|---|---|---|---|---|---|---|---|---|
| Task 1 | Intrinsic Shannon Entropy for Second IMF of X-coordinate (67.86) | StdDev of In Air jerk in X direction (66.75) | GeoMean of In Air jerk in Y-direction (72.50) | Median of In-Air Acceleration in Y-direction (69.71) | Intrinsic third-order Renyi Entropy for Second IMF of X-coordinate (72.44) | 40% Trimmed Mean of In-Air Velocity in Y-direction (75.83) | Median of In-Air Acceleration in Y-direction (90.00) | Third Moment of In-Air Acceleration in Y-direction (84.00) | 90th Percentile of In-Air velocity in X-direction (72.00) |
| Task 2 | Arithmetic Mean of In Air Acceleration in X direction (67.60) | RobustRange of In Air Velocity in X direction (71.00) | Kurto of On Surface Velocity (62.20) | 1st Percentile of Pressure rate (67.14) | RobustRange of In Air jerk in Y direction (78.66) | Third Moment of In-Air jerk (74.17) | 1st Percentile of Pressure rate (70.00) | Arithmetic Mean of In-Air Acceleration (**91.99**) | 1st Percentile of Pressure rate (61.00) |
| Task 3 | First Percentile of Pressure Rate (44.79) | Relative NCV In Air (61.50) | 20nd Percentile of On Surface velocity in X-direction (70.25) | 1st Percentile of Pressure rate (64.57) | 1st Percentile of Pressure rate (65.77) | 1st Percentile of Pressure rate (75.00) | Relative NCA In-Air (75.00) | Geometirc Mean of In-Air Jerk (70.80) | Geometirc Mean of In-Air velocity in Y-direction (68.50) |
| Task 4 | 30th Percentile of x component of velocity on Surface (67.73) | First Percentiles of Pressure Rate (40.00) | StdDev of OnSurface Acceleration in X-direction (69.75) | 1st Percentile of Pressure Rate (68.88) | Range of In-Air Velocity in Y-direction (72.88) | Intrinsic third order Renyi Entropy for Second IMF (73.33) | RobustRange of Pressure Rate (80.00) | Range of On-SurfaceVelocity (61.60) | 1st Percentile of Pressure rate (58.00) |
| Task 5 | 30% Trimmed Mean of On Surface Acceleration in X direction (62.26) | Intrinsic second order Renyi Entropy for First IMF of Y-coordinate (**72.50**) | Mode of On Surface Velocity in X-direction (69.25) | 1st Percentile of Pressure Rate (66.57) | 1st Percentile of Pressure rate (66.88) | 1st Percentile of Pressure rate (69.50) | 1st Percentile of Pressure rate (75.00) | Intrinsic second order Renyi Entropy for First IMF in Y-direction (60.00) | 1st Percentile of Pressure rate (63.00) |
| Task 6 | Relative NCV on surface (61.20) | Relative NCV On Surface (64.00) | 90th Percentile of On Surface Velocity in X-direction (77.00) | 1st Percentile of Pressure Rate (68.85) | 20% Trimmed Mean of On Surface Velocity in X-direction (66.22) | 1st Percentile of Pressure rate (64.50) | Relative NCA In-Air (**96.00**) | Kurtosis of Off-surface velocity in y-direction (73.20) | RobustRange of In-Air Jerk (**79.00**) |
| Task 7 | SNR of ICE of x-coordinate (**77.20**) | SNR of ICE of x-coordinate (62.25) | SNR of ICE of X-coordinate (**83.25**) | 95th Percentile of In-air Velocity in X-direction (74.28) | SNR of ICE of X-coordinate (**79.33**) | SNR of CE of y-coordinate (**89.50**) | 1st Percentile of Pressure rate (77.00) | SNR of ITKE of x-coordinate (71.20) | 1st Percentile of Pressure rate (65.00) |

TABLE 4
MANN-WHITNEY U TEST RESULTS (4-WAY)

| Task Number | # features passed (*OF*) | # features passed (*YF*) | # features passed (*OM*) | # features passed (*YM*) |
|---|---|---|---|---|
| Task 1 | 13 | 14 | 73 | 9 |
| Task 2 | 38 | 3 | 113 | 2 |
| Task 3 | 7 | 5 | 4 | 2 |
| Task 4 | 9 | 5 | 4 | 1 |
| Task 5 | 3 | 3 | 5 | 5 |
| Task 6 | 1 | 10 | 31 | 31 |
| Task 7 | 98 | 3 | 20 | 20 |

(OF- Old Female; YF- Young Female; OM- Old Male; YM- Young Male)

Table 3 shows the number of features in each category of classification for each handwriting task which passed the Mann-Whitney U Test. The summation of each column in Table 3 gives the maximum feature dimension for that category. A large number of features corresponding to Task 7 passed Mann Whitney U Test suggesting that these features play a dominant role in the classification of Parkinson's and healthy individuals in both male and female participants.

Mann Whitney results for a 4-way classification are presented separately in Table 4. It also aids in gaining an insight into the kind of task that is more prominent in PD classification. We discuss it in detail in the Results as well as in the Discussion section.

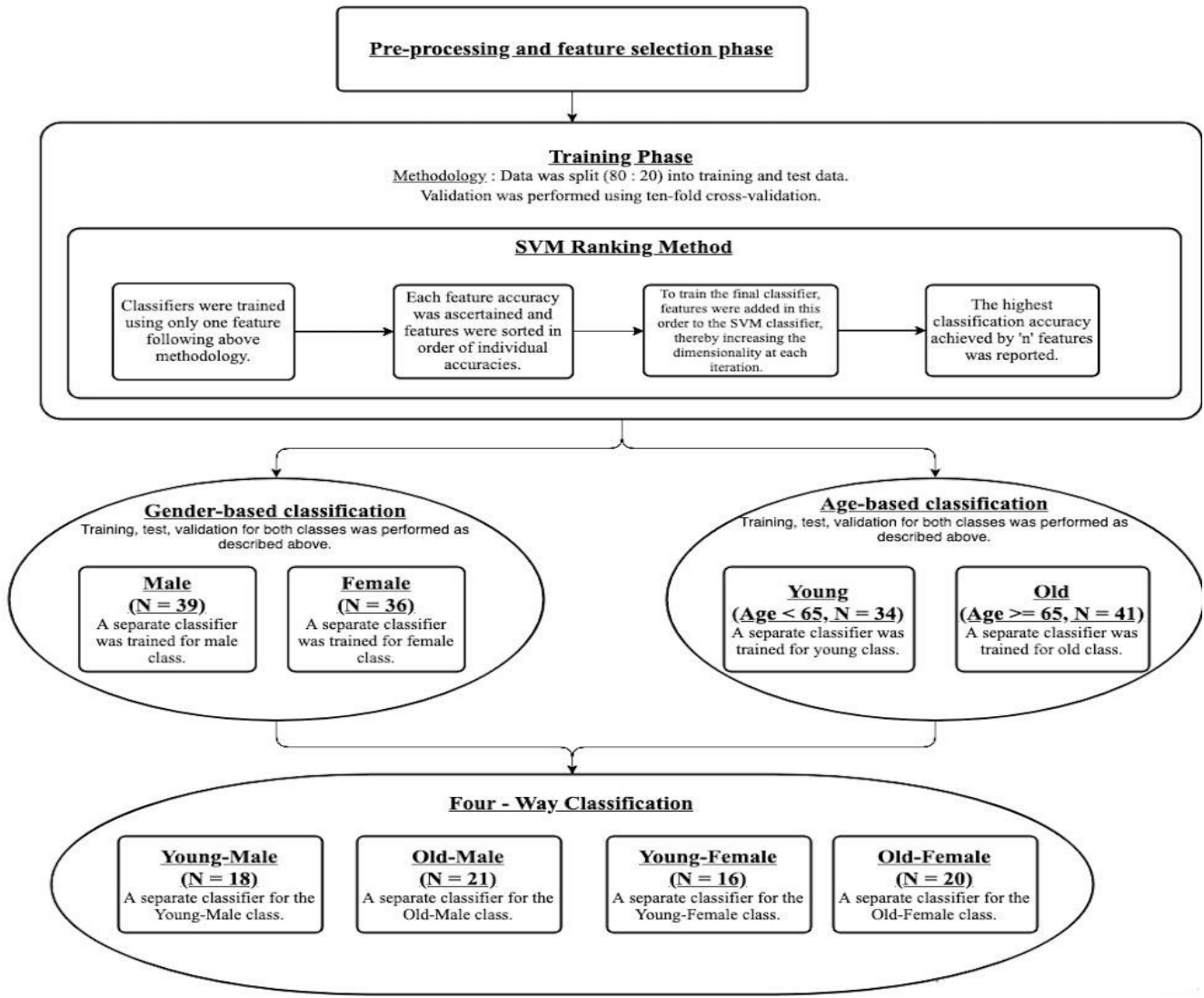

Fig. 3 Flowchart for classification in different categories

TABLE 6
NUMBER OF FEATURES OF EACH TYPE IN TOP10 SELECTED FEATURES

| Category | Number of Kinematic Features in Top10 | Number of Entropic Features in Top10 | Number of Energetic Features in Top10 |
|---|---|---|---|
| Male | 7 | 2 | 1 |
| Female | 2 | 5 | 3 |
| Old | 8 | 0 | 2 |
| Young | 10 | 0 | 0 |

The single feature with highest accuracy for each task is presented in Table 5. Since individual feature accuracies are incorporated in the feature selection phase, the validation set was used to estimate the values reported in Table 5. The best individual feature accuracy for female class is 83.25% which is much higher than for the combined category classification accuracy 77.20%. We observe that the best individual feature accuracy for the Young Female class is 96%, while it is 79% for the Young Male.

Similarly, the old-age classifier (79.33%) outperformed combined classifier with an improvement of nearly 3% accuracy for an individual feature (SNR of ICE of x coordinate) of a given task 7.

Table 6 enlists the number of features belonging to Kinematic, Entropic, and Energetic type from the top 10 features for each category found through descending SVM Ranking approach. The top-performing features for the male class were mainly kinematic, while for the female class, they were entropic and energetic features. The top-performing features for old class and young class were predominantly kinematic. Performance of Random Order Ranking was compared with Descending Order Ranking Approach, as shown in Fig.4 (a) and Fig.4 (b). Since the Descending Order SVM ranking approach remained above the Random Order approach for most of the performance curve, we prefer the former in this paper.

Table 7 reports performance parameters for descending order SVM ranking. It shows the accuracy, precision, and recall for sex and age-dependent categorization on the test set. The maximum accuracy was obtained for N=4 features for SVM Ranking method in descending order while the maximum accuracy obtained for Random approach was achieved for N=3 features.

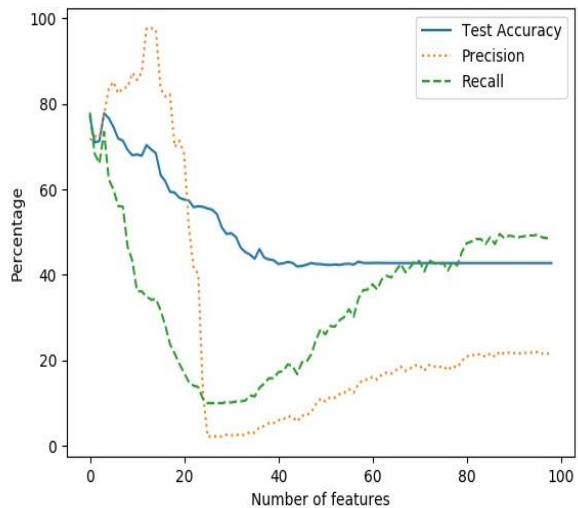

Fig.4 (a) SVM Ranking approach with descending order feature addition

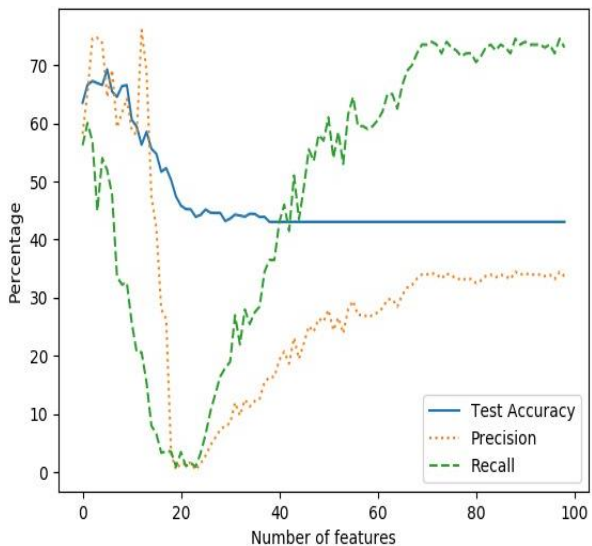

Fig.4 (b) Random order feature addition

**TABLE 7**
OVERALL PERFORMANCE OF EACH CLASS

| Class | $P_{acc}$(%) | $P_{pre}$(%) | $P_{rec}$(%) |
|---|---|---|---|
| Combined | 75.76 | 97.72 | 81.02 |
| Male | 74.00 | 83.50 | 65.54 |
| Female | 83.75 | 94.40 | 85.07 |
| Old | 79.55 | 64.52 | 77.27 |
| Young | 70.62 | 84.26 | 80.46 |

Fig.5 shows the classification accuracy for sex-specific (male and female) and age-dependent (young and old) classification model. The female-specific classification model significantly outperformed (p<0.05) the generalized model with an improvement of nearly 8% in classification accuracy. Similarly, the information of age for an old age-dependent classification model significantly improved (p<0.05) the classification accuracy by a margin of nearly 4% in comparison to the generalized classification scheme.

Table 8 shows the performance parameter values for the 4-way classification, where age and sex information both are fed to the classifier which led to four classifiers as follows – Young Male, Young Female, Old Male, and Old Female. As mentioned earlier, increasing the number of groups reduces the number of samples for training and testing in each group, which is why achieving 96.25% accuracy on Young Female class might not be statistically significant. We further discuss the class imbalance and insufficient data issue in the next section.

**TABLE 8**
OVERALL PERFORMANCE OF EACH CLASS (4-WAY)

| Class | $P_{acc}$(%) | $P_{pre}$(%) | $P_{rec}$(%) |
|---|---|---|---|
| Old Female | 74.75 | 14.5 | 12.83 |
| Young Female | 96.25 | 99.00 | 98.75 |
| Old Male | 84.99 | 73.13 | 78.11 |
| Young Male | 69.00 | 71.66 | 92.41 |

## VII. DISCUSSION

Agreeing to our hypothesis, the sex-specific classifier outperformed the generalized classifier. The present results are in-line to previous recent work where female-dependent classifier dominated over male-dependent and generalized classifier. The previous work has shown the improved accuracy in movement disorder recognition using age and gender grouping [39].

The presented accuracy in our work for Female-dependent model is better than reported existing pathological examination determination [40]. However, it has to be noted that the present work just classifies healthy individuals from individuals with PD. The extension of current work where individuals with PD need to be distinguished from individuals with a movement deficit would be of great importance in the future. The current work suggests that Sex and Age-based prior information may be crucial for such future applications.

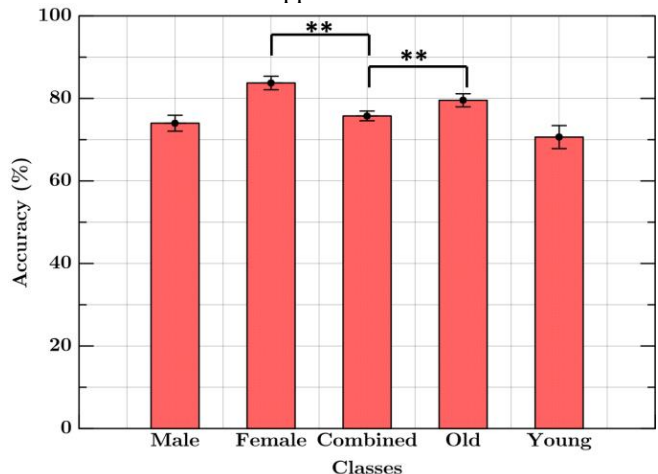

**Fig.5** Classification Accuracy for different category of classification

Further, Tables 1, 3, and 5 elucidate the specificity of each category for a given feature/task can be biologically explained from neurological and physiological differences between sex and age groups [42]. As shown in Table 4, the number of passed features, which were able to discriminate healthy and Parkinson's individuals, depending upon the category of classification. Similarly, Table 5 shows that the capacity of each task in classification of Parkinson's and healthy individuals depends on the category. For example, the individual feature for Task 3, which performs best in classification is different for male and female. Similarly, different age-specific task and features were observed for classification. It provides a concrete foundation to pursue the PD classification task through sex and age-based grouping. The dimensionality of the input feature space was reduced significantly after the Mann-Whitney Test as only a few features were able to pass the Mann-Whitney U test.

It can be seen from Table 3 that Task 7 provides the most number of passed features for classification. It is intuitive as Task 7 involves the subjects to write more complex words as compared to other tasks. It can also be inferred from Table 5 that features corresponding to Task 3 do not perform well on Combined as well they do for Male/Female Class. This analysis provides an excellent platform to work on designing handwriting tasks which have better Capacity than existing functions for each class. From the study conducted by [49], the best performing tasks were 3, 6, 8 (corresponding to 2,5,7 in Table 5). In agreement with this, we observe that task 2 provides the best performing individual feature(highlighted in bold in Table 5) in the Old Male class, task 5 achieves the same for the Male class, while task 7 achieved the same for four different classes namely- Combined, Female, Old and Old Female class. However, we also notice that task 6 provides best performing individual feature for two different classes namely- Young Female and Young Male. Although, it must be noted that the determination of best performing tasks in [49] differs significantly from ours.

Table 6 reveals that the dominant features which can be used for determining PD in males and females. It is possible because of their inherent neurological differences [43, 44, 45]. Similarly, we show that certain features should differ for subjects above 65 (considered old) and those below 65 (considered young). Finally, feature addition did not improve further accuracy as shown in Fig.4 (a) and Fig.4 (b), instead deteriorate, which could be due to the increasing dimensional feature space without increasing the training example proportionally leading to overfitting.

Table 7 shows that the accuracy of our model has increased immensely to $P_{acc}$= 83.75%, $P_{acc}$= 79.55% in Female, Old class respectively, which is higher than Combined class. This helps to understand the impact of age and sex in the classification process. It should be noted that the sex-based dataset is age-balanced(mean values of females and males are close to each other), and the age-based dataset is sex-balanced(number of males and females in young and old class is similar) as indicated in Table 2. This implies that the incremental improvement in the performance for a class(age/sex) over the combined class is independent of the other class(sex/age). However, it must be noted that since the age-based dataset suffers from the problem of class imbalance, we also report precision and recall scores. The corresponding Precision and Recall for the female class is also higher than for Combined Class. Since we are trying to create a preliminary diagnostic test for PD, higher recall scores have greater clinical value, as subjects with PD are indeed classified as having PD in the model with high recall. The high recall, along with lower precision score shows that the model can detect PD but misclassifies some healthy subjects as having PD. The classification accuracy of Male has not improved when compared to the Combined level, which suggests that the selected writing tasks are not good enough to be used for classification into HC and PD in this case.

**TABLE 9**
PERFORMANCE COMPARISON OF PARKINSON'S DISEASE DETECTION SYSTEM

| Study | Features | Classifier | Dataset | Accuracy(%) |
|---|---|---|---|---|
| Drotár et al. 2015 [37] | Entropic, Energetic | SVM-RBF | PaHaW | 88.1 |
| Impedovo et al. 2018 [46] | Kinematic, Entropic, Energetic, Pressure, Extended Velocity-Based signals | SVM-Linear | | 93.79 |
| Cioppa et al. 2019 [47] | Kinematic | Cartesian Genetic Programming Approach | | 76.6 |
| Naseer et al. 2019 [48] | Fine tuned Image Net features | AlexNet | | 98.28 |
| Proposed System | Kinematic, Entropic, Energetic, Pressure | SVM-RBF | | 75.76 |
| Proposed System | Kinematic, Entropic, Energetic, Pressure | SVM-RBF(Female) | | 83.75 |

The 4-way classification was also performed for which the results are tabulated in Table 4 and Table 8. It is important to note that 4-way classification suffers from class imbalance and data insufficiency. As clear from Table 2, the number of subjects on which training was performed significantly reduces during Train Test split. Additionally, the number of healthy control subjects is substantially different from the number having PD causing class imbalance. In the future, a thorough analysis using more number of training samples as well as balanced data for each class can be performed for more insights. The diagnosis of Parkinson's in early-stage can improve the patient's quality of life as well as the cost of treatment [45]. In the present study, nearly 80 % of PD patients were at early stages (UPDRS < 2.5) with a distribution of UPDRS score as following: 1 (n =5), 2 (n = 18), 2.5 (n = 6), 3 (n =5), 4 (n = 2), and 5 (n = 1). It shows the potential of the present approach in the identification of PD patients at early stages; however, a future study with a large number of patients may further improve the model for higher identification accuracy.

We also analyze and compare the performance of the proposed system with existing Parkinson's classification systems on PaHaW as the target dataset. Table 9 illustrates the performance values for various state-of-the-art detection systems along with their brief descriptions. It must be noted that the other authors have reported the accuracy values as the average of the scores obtained by stratified cross-validation. However, we report our results over an independent test set, as described in the SVM Ranking subsection, which leads to lower performance values.

## VIII. CONCLUSION

We have shown that the proposed scheme can be used for diagnosis of PD with classification accuracy over 80% using sex-specific and age-dependent distinction. The division into sex and age provided insights into the differentiability of a feature and writing task to serve as a marker for PD. Further, as age and sex determination do not require any instrumentation or computation, it does not add to any further needed resources like in other pathological methods. The order in which the SVM ranked features are trained deemed to be an essential factor while calculating accuracy, Precision, and Recall. We observed that the accuracy of the model reduces if the features are added in random order rather than decreasing order. We intended to introduce the idea of 4-way classification in this work. A more rigorous analysis can be performed on the 4-way division with more as well as balanced data. Performing Sampling (Data Augmentation) on the given data could be one of the option to address insufficient data issue; however, an alternative is needed to answer the class imbalance problem.